\pdfoutput=1

\PassOptionsToPackage{dvipsnames}{xcolor}
\documentclass[11pt]{article}

\usepackage[final]{acl}

\usepackage{times}
\usepackage{latexsym}
\usepackage{bm}

\usepackage[T1]{fontenc}

\usepackage[utf8]{inputenc}

\usepackage{microtype}

\usepackage{inconsolata}

\usepackage{graphicx}
\usepackage{url}            
\usepackage{booktabs}       
\usepackage{amsfonts}       
\usepackage{nicefrac}       
\usepackage{microtype}      
\usepackage{xcolor}         
\usepackage{tcolorbox}
\usepackage{amsmath}
\usepackage{soul} 
\usepackage{multirow}
\usepackage{booktabs}
\usepackage{adjustbox}
\usepackage{hhline}
\usepackage{tabularx} 
\usepackage{subcaption}
\usepackage{enumitem}
\usepackage{wrapfig}

\title{Deciphering the Factors Influencing the Efficacy of Chain-of-Thought: Probability, Memorization, and Noisy Reasoning}

\author{
 \textbf{Akshara Prabhakar,\textsuperscript{1}}
 \textbf{Thomas L. Griffiths,\textsuperscript{1,2}}
 \textbf{R. Thomas McCoy\textsuperscript{3,4 }\Thanks{Work begun while at Princeton University.}}
\\
\\
 \textsuperscript{1}Department of Computer Science, Princeton University\\
 \textsuperscript{2}Department of Psychology, Princeton University\\
 \textsuperscript{3}Department of Linguistics, Yale University
 \\
 \textsuperscript{4}Wu Tsai Institute, Yale University
\\
 \small{
   \textbf{Correspondence:} \href{mailto:akshblr555@gmail.com}{akshblr555@gmail.com}
 }
}

\definecolor{darkgreen}{RGB}{0, 100, 0}
\definecolor{lightorange}{RGB}{255, 187, 51}
\definecolor{darkpink}{rgb}{0.4, 0.1, 0.2}
\newcommand{\q}[1]{``#1''}

\begin{document}
\maketitle

\begin{abstract}
  Chain-of-Thought (CoT) prompting has been shown to enhance the multi-step reasoning capabilities of Large Language Models (LLMs). 
However, 
debates persist about whether LLMs exhibit \textit{abstract generalization} or rely on \textit{shallow heuristics} when given CoT prompts.
To understand the factors influencing CoT reasoning we provide a detailed case study of the symbolic reasoning task of decoding shift ciphers \cite{ANDRESS201469}, where letters are shifted forward some number of steps in the alphabet. 
We analyze the pattern of results produced by three LLMs---GPT-4, Claude 3, and Llama 3.1---performing this task using CoT prompting.
By focusing on a single relatively simple task, we are able to identify three factors that systematically affect CoT performance: the probability of the task's expected output (probability), what the model has implicitly learned during pre-training (memorization), and the number of intermediate operations involved in reasoning (noisy reasoning). We show that these factors can drastically influence task accuracy across all three LLMs; e.g., when tested with GPT-4, varying the output's probability of occurrence shifts accuracy from $26\%$ to $70\%$. Overall, we conclude that CoT prompting performance reflects both memorization and a probabilistic version of genuine reasoning.\footnote[2]{Code and data are available at \url{https://github.com/aksh555/deciphering_cot}.}

\end{abstract}

\section{Introduction}
\label{sec:introduction}
Reasoning, one of the key aspects of human intelligence, is the process of thinking about something logically and systematically using evidence and past experiences to make a decision \cite{wason1968reasoning,wason1972psychology,fagin2004reasoning}. The impressive performance of Large Language Models (LLMs) across a wide range of tasks has spurred extensive research into their reasoning capabilities \cite{huang2022towards,qiao2022reasoning}. It remains unclear whether the behavior of these systems is based on true reasoning or on shallow heuristics. 
Some results provide evidence that LLMs are able to reason \cite{suzgun2022challenging,dasgupta2023language,saparov2022language}, while others show that they still struggle on tasks that humans can easily solve via reasoning  \cite{han2022folio,valmeekam2023planbench,mccoy2023embers,razeghi2022impact,cao2023retentive}.

The Chain-of-Thought \cite[CoT;][]{cot} prompting strategy has played a significant role in this debate. CoT involves prompting an LLM to generate a sequence of intermediate reasoning steps before producing the final answer, given some in-context exemplar(s) of how to break the task  into steps. 
CoT and its several variants \cite{kojima2023large,zhou2022large,wang2023selfconsistency} have been shown to substantially improve performance over standard prompting.
Recent works have tried to identify which aspects of the demonstration contribute to CoT's enhanced performance \cite{huang2022towards,madaan2022text,jin2024impact}, typically relying on assessing performance across a wide range of  tasks. 

\begin{figure*}[t]
    \centering
\includegraphics[width=0.95\textwidth]{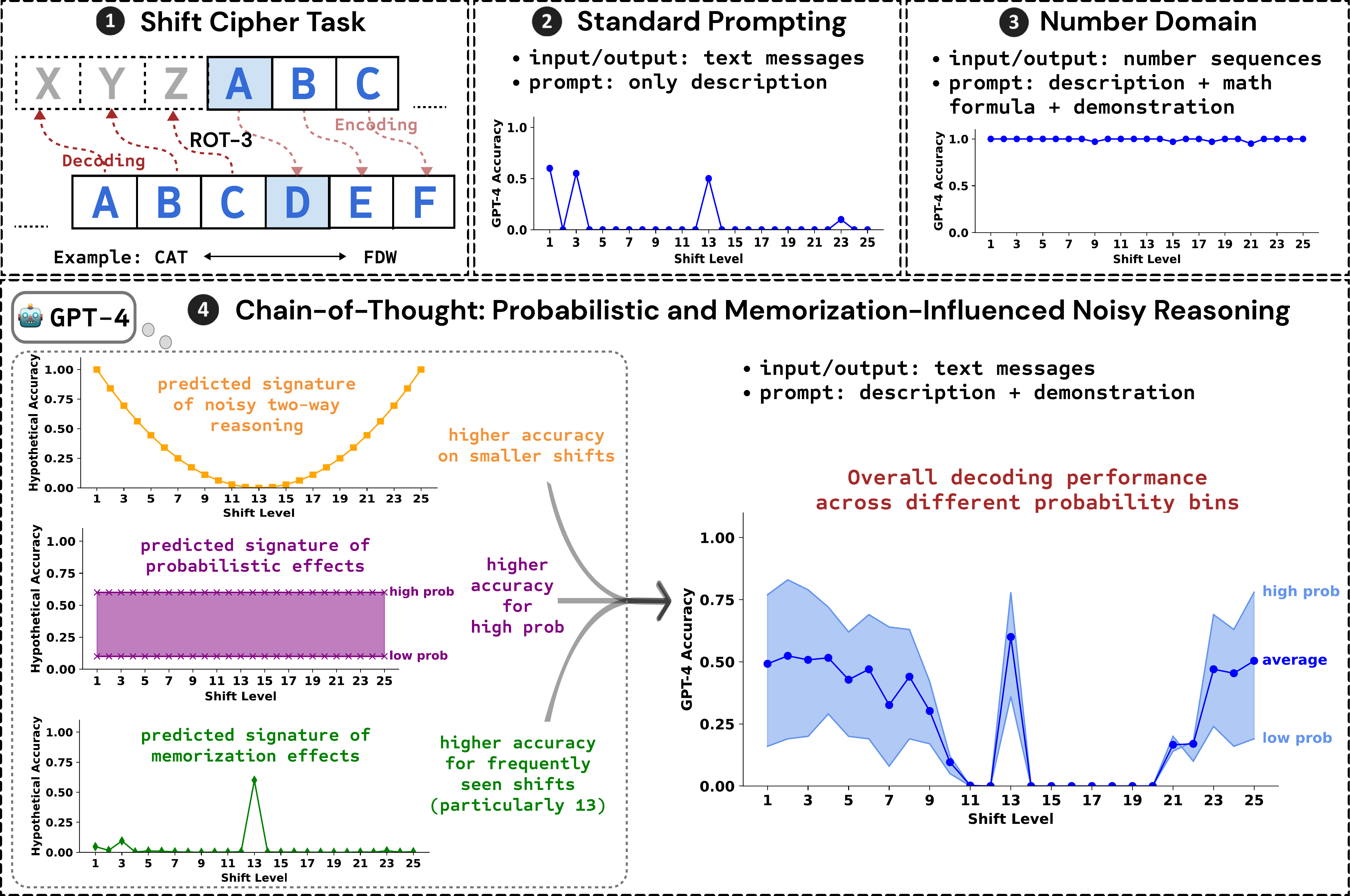}
    \caption{Overview. (1) Task: We have LLMs decode messages written in a shift cipher, in which each letter is shifted a fixed number of positions forward in the alphabet. (2) With standard prompting, GPT-4 performs poorly across most shift levels. (3) However, GPT-4 scores nearly perfectly on an isomorphic task based on numbers rather than letters. (4) With CoT prompting, GPT-4 adopts probabilistic and memorization-influenced noisy reasoning. That is, its performance (right) combines the trends we have hypothesized for each of the three factors on the left.
    }
    \label{fig:preview}
\vspace{-1pt}
\end{figure*}

In this work, we take a different approach: we present an extensive case study on a single task that allows us to disentangle reasoning from memorization. The task we selected is solving shift ciphers, a simple type of code in which each letter is shifted forward a certain number of positions in the alphabet (\autoref{fig:preview}, panel 1). We choose this task because it allows us to independently manipulate several factors that could be relevant for characterizing how LLMs solve reasoning tasks when prompted with CoT: difficulty, frequency, and answer probability. 

Our results suggest that CoT performance reflects three factors: probability, memorization, and noisy reasoning. 
First, the accuracy of CoT is affected by the probability of the correct output, with more probable outputs resulting in a stronger effect of CoT. Second, performance is higher when memorization is possible, as indicated by the frequency of encountering different shift cipher variants during pre-training. 
The effects of probability and memorization show that CoT performance is not fully systematic abstract reasoning.
Nonetheless, CoT performance is not solely driven by superficial heuristics: it also shows some hallmarks of true reasoning---albeit a noisy version of true reasoning, in which the error rate increases along with task difficulty (where we quantify task difficulty by the number of implicit reasoning steps involved).
In addition, we find evidence that the effect of CoT fundamentally depends on generating sequences of words that increase the probability of the correct answer when conditioned upon; as long as this is the case, CoT can thus succeed even when the demonstrations in the prompt are invalid. 
In the ongoing debate about whether LLMs reason or memorize \cite{feldman2020does,zhang2023counterfactual,magar2022data,srivastava2024functional,antoniades2024generalization}, our results thus support a reasonable middle-ground: \textit{LLM behavior displays aspects of both memorization and reasoning, and also reflects the probabilistic origins of these models.}


\section{Related Work}
\label{sec:related}
\paragraph{In-Context Learning in Language Models.}
It has been argued that LLMs can \textit{learn} a task purely from demonstrations given in their context without any additional training \cite{NEURIPS2020_1457c0d6}, a phenomenon known as in-context learning (ICL). There have been many investigations into how ICL operates. Theoretical frameworks have modeled the pretraining data as a mixture of Hidden Markov Models \cite{xie2022explanation} and have argued that ICL is the result of implicit Bayesian inference \cite{zhang2023and}, an argument supported with evidence from synthetic data and tasks \cite{chiang2024understanding}. The emergence of ICL has been attributed to factors including data distributional properties \cite{NEURIPS2022_77c6ccac}, pretraining term frequencies \cite{razeghi2022impact}, and the creation of task vectors \cite{hendel-etal-2023-context}. However, the extent to which ICL is \textit{true learning} is unclear \cite{pan-etal-2023-context,min2022rethinking}. For example, \citet{kossen2024incontext} shows that ICL relies on in-context label information but cannot fully overcome preferences acquired during pre-training, which is evidence against the view that ICL is true learning.

\paragraph{Understanding CoT. }
Theoretical arguments have been formulated about how CoT improves the expressivity  \cite{NEURIPS2023_dfc310e8,li2024chain,merrill2024expressivepowertransformerschain} and sample complexity \cite{li2023dissecting} of ICL. 
Empirical studies have shown that CoT performance can be dramatically influenced by many features of the CoT prompt \cite{madaan2022text,wu2023analyzing,jin2024impact}; for example, the relevance and ordering of reasoning fragments is more important than their accuracy \cite{wang-etal-2023-towards,wang2023selfconsistency,ye2022the}, and minor input perturbations can substantially bias models' answers \cite{turpin2024language}, suggesting that the models lack general reasoning abilities \cite{stechly2024chain}. 


Our high-level goal in this work is to characterize what \textit{type} of reasoning happens in LLMs when they are prompted with CoT prompting: to what extent is CoT performance driven by abstract reasoning vs.\ simple heuristics such as memorization? 
This question also implicitly underlies many of the papers discussed above, but directly focusing on this question leads us to investigate probability, frequency, and difficulty (see \autoref{ss:reasoning_formulation})---a different set of factors than those studied in prior work.

\section{Approach}
\label{sec:methods}


One challenge of evaluating the role of memorization and reasoning in the performance of LLMs is that these models are typically evaluated on a wide range of complex reasoning tasks, whose variety and complexity can obscure the factors that drive performance. By contrast, we propose to tease apart the factors behind the efficacy of CoT prompting by focusing on a single relatively simple task: deciphering text encoded with a shift cipher. 

Encoding a message with a shift cipher involves replacing every letter with another letter that is some fixed number of positions (called $shift\_level$) forward in the alphabet; decoding is the reverse (shifting backward) as shown in \autoref{fig:preview}. These are also known as rotation ciphers since they rotate the alphabet forward some number of steps, and they are given the name \texttt{rot-$k$} where $k$ corresponds to $shift\_level$. For example, given the test word \q{FDW} and that  \texttt{rot-3} encryption has been used ($shift\_level = 3$), decoding involves shifting every letter $3$ steps backward---i.e., \textit{F $\rightarrow$ C}, \textit{D $\rightarrow$ A}, and \textit{W $\rightarrow$ T} to obtain \q{CAT} as the output. In our experiments, we give an LLM a single word encoded with a shift cipher and ask it to decode this text to recover the original word.

\subsection{Motivation for using shift ciphers}
\label{ss:motivation}
Our main reason for using shift ciphers is 
because they involve a sharp dissociation between task complexity 
and task frequency (a key factor in memorization). The complexity of the decipherment task is determined by the shift level---ciphers that require more intermediate steps are more complex. Different shift levels also vary in their frequency in internet text \cite{mccoy2023embers}, and hence in the training data of large language models. Specifically, \texttt{rot-13} is widely used in internet forums to conceal text such as puzzle solutions and spoilers, and \texttt{rot-3} and \texttt{rot-1} commonly appear in tutorials on decipherment (\texttt{rot-3} is also known as the Caesar cipher, having apparently been used by the eponymous Caesar to encrypt his messages).
In addition, shift ciphers facilitate investigation of the effect of probability because the correct answer can be any string, allowing us to modulate the probability of that string easily. 
Further, the systematic nature of the task makes it easy to generate examples and to verify correctness. Finally, decoding each letter in the message is an independent step, allowing us to easily analyze these individual steps. 

\citet{mccoy2023embers} previously evaluated GPT models on shift ciphers, focusing on standard prompting along with some initial results using CoT. They study the effect of only probability and memorization, while we conduct a more extensive investigation into LLM behavior when prompted with CoT by additionally studying the influence of complexity and analyzing more models. Importantly, we add nuance to their findings by arguing for a middle-ground viewpoint that acknowledges the LLM weaknesses identified by \citet{mccoy2023embers} but also brings in novel observations that highlight the hallmarks of true reasoning that are present in these systems. 

\subsection{The effect of CoT on shift ciphers}
\label{sec:exp_details}
\paragraph{Data.}
\label{ss:word_dataset}

We constructed a dataset comprising 7-letter words having exactly 2 tokens (measured using the tokenizer used by GPT-4) to control for confounding factors relating to tokenization. We found all 3-letter and 4-letter tokens from the lowercase English alphabet and formed words by considering possible combinations of 3-letter word-initial tokens followed by  4-letter non-word-initial tokens.
Following \citet{mccoy2023embers}, we compute the log probability as the log probability that GPT-2 \cite{radford2019language} assigns to the sentence `\texttt{The word is "WORD"}', minus the log probability that it assigns to `\texttt{The word is "}'; thus, this yields the log probability assigned to just the word and the following quotation mark in the context of `\texttt{The word is "}'. The closing quotation mark is included because it indicates the end of the word.  The words were scored by their log probability and arranged in descending order. Subsequently, five bins 
were formed by selecting equidistant log probability values as centers, with \texttt{bin1} having the highest probability and \texttt{bin5} having the lowest probability. We manually checked the words in this dataset and filtered them to ensure there were no inappropriate words used to obtain 150 words for each bin. We partitioned the 150 examples into two subsets: a subset containing 100 words used to evaluate GPT-4, and a subset containing 50 words used to evaluate logistic regression models that were fitted to GPT-4's performance on the 100-word subset. We prepared the inputs for the models by producing the shift-cipher-encoded versions of the words from the 5 probability bins across 25 shift levels (1 to 25). We ran all evaluations a single time; the accuracies that we report are accuracies over these 100-example sets.

We then assessed performance on this dataset using a variety of different prompts:
\begin{itemize}[leftmargin=1em,noitemsep,topsep=0pt]
    \item[--]\textbf{Standard.} This is a prompt with just the description of the task and demonstration but no reasoning steps (\autoref{fig:standard_prompt}).
\begin{figure}
\centering
\begin{tcolorbox}[
    width=0.5\textwidth,
    fontupper=\ttfamily,
    left=0.5mm,
    right=0.5mm,
    top=0.5mm,
    bottom=0.5mm,
    title=Standard,
    center title,
    fonttitle=\tiny\ttfamily]
\tiny
\textcolor{blue}{Rot-13 is a cipher in which each letter is shifted 13 positions forward in the alphabet.} \textcolor{orange}{For example, here is a message written in rot-13 along with the original text that it was created from:\\
Rot-13 text: "fgnl"\\
Original text: "stay"}\\

Decode this message to produce the original text:

Rot-13 text: <test\_input>
\end{tcolorbox}
\caption{Standard prompt having just the \textcolor{blue}{description} and \textcolor{orange}{demonstration}.}
\label{fig:standard_prompt}
\end{figure}
 
    \item[--] \textbf{Text-CoT.} This prompt encourages the model to decode a message one letter at a time (\autoref{fig:text_cot_prompt}). We chose this way of framing the CoT prompt following \citet{mccoy2023embers}, who tried several variants and found this to be the best. To get a reasoning step correct, the model must have learned the alphabet during pre-training.
    \item[--]\textbf{Math-CoT.}
The prompt (Appendix~\ref{appx:prompts} \autoref{fig:math_cot_prompt}) encourages a reasoning pipeline that involves translating each letter to a number, performing the shift by applying arithmetic to this number, then converting the result back to a letter. The prompt also specifies the mapping between letters and positions, eliminating the need for the model to have internalized the positions of the letters in the alphabet.\begin{figure}[t]
        \centering
\begin{tcolorbox}[
    width=.47\textwidth,
    fontupper=\ttfamily,
    left=0.5mm,
    right=0.5mm,
    top=0.5mm,
    bottom=0.5mm,
    title=Text-CoT,
    center title,
    fonttitle=\tiny\ttfamily]
\tiny
\textcolor{blue}{Rot-13 is a cipher in which each letter is shifted 13 positions forward in the alphabet.} \textcolor{orange}{For example, here is a message written in rot-13:\\
Rot-13 text: "fgnl"}\\

\textcolor{orange}{To decode this message, we shift each letter 13 positions backward:}\\
\textcolor{purple}{1. f -> s\\
2. g -> t\\
3. n -> a\\
4. l -> y}\\
\textcolor{orange}{Therefore, the original text is: "stay"}\\

Here is another message in rot-13. Decode this message one letter at a time. On the last line, write the words "Original text:" followed by the decoded message:\\
Rot-13 text: <test\_input>
\end{tcolorbox}

\caption{Text-CoT prompt consisting of a \textcolor{blue}{description} and \textcolor{orange}{demonstration} that includes several \textcolor{purple}{reasoning steps}.}
\label{fig:text_cot_prompt}
\end{figure}

    \item[--]\textbf{Number-sequence CoT (Number-CoT).} This prompt (Appendix \ref{appx:prompts} \autoref{fig:math_only_prompt}) makes use of an alternative task that is isomorphic to shift ciphers but based in the number domain---the input and output are number sequences instead of letter sequences. Reasoning involves applying arithmetic to the input elements in the number sequence to get a corresponding output sequence. 
\end{itemize}

\noindent
We ran experiments using both open and closed source models: GPT-4 (\texttt{gpt-4-0613}) \cite{openai2023gpt4}, Claude 3 (\texttt{claude-3-opus-20240229}) \cite{claude}, and \texttt{Llama-3.1-405B-Instruct} \cite{dubey2024llama3herdmodels}.
The reason for using such strong models is that their shift cipher performance is significantly improved by prompting with chain of thought. Additionally, this helps us to control several sources of extraneous errors making it easier to focus on the task itself and isolate the factors affecting CoT: It ensures that the format of the demonstration is closely followed, and that copy errors (errors in copying information from the prompt such as letters from encoded text and letter-position mappings) are rare.
We set \texttt{temperature} to $0$ and \texttt{max\_new\_tokens} to $200$. 

\autoref{fig:preview} provides some initial results for GPT-4. Using standard prompts,
GPT-4 gets zero accuracy across most shift levels, but it improves substantially (to an average accuracy of 32\%) when Text-CoT is used; this result replicates the finding in \citet{mccoy2023embers} that CoT is helpful for shift ciphers 
but still remains far from perfect. However, with Number-CoT, GPT-4's performance becomes nearly perfect (more details are in Appendix \ref{appx:prompts}).

These results paint CoT prompting in a puzzling light. Prompting with Number-CoT showed that GPT-4 has the core reasoning abilities that would be needed to decode shift ciphers nearly perfectly. Thus, if CoT prompting led to symbolic reasoning, GPT-4 would score perfectly. The fact that it does not shows that CoT reasoning is not pure symbolic reasoning. Nonetheless, it is also clear that CoT does substantially improve over standard prompting, so it is unlikely that CoT reasoning can be explained away as simple memorization.
If CoT reasoning is neither simple memorization nor pure symbolic reasoning, what is it? This question motivates our experiments in the next section.

\section{Disentangling the
factors influencing CoT performance}
\label{ss:reasoning_formulation}
We consider four types of reasoning processes that LLMs might be adopting.


\begin{enumerate}[label=(\alph*),noitemsep,topsep=0pt,leftmargin=2em]
\item \textbf{Symbolic reasoning} is the use of discrete, deterministic inference rules. Shift ciphers can be perfectly decoded with a simple symbolic algorithm, so a system using fully systematic reasoning should attain 100\% accuracy.
\item \textbf{Noisy reasoning} is like symbolic reasoning but with the addition of noise that introduces some possibility of each intermediate operation in a reasoning step being wrong. Thus, if the system uses noisy reasoning, we should see accuracy decrease as we increase the number of operations that need to be performed. Shift ciphers let us test this possibility: by varying $shift\_level$, we can modulate the number of operations that need to be performed in every reasoning step and observe if accuracy varies accordingly.
\item \textbf{Memorization} is a strategy in which a system memorizes the tasks it has encountered in pre-training but does not generalize to new tasks. If memorization is all that LLMs do, we should see higher performance in the cases that are frequently encountered during pre-training than the ones that are not. \citet{mccoy2023embers} show that $13$ is by far the most common shift level in natural corpora because this shift level (sometimes called \texttt{rot-13}) is popular in some online communities. Thus, a hallmark of memorization would be much higher accuracy at $13$ than other shift levels.
\item \textbf{Probabilistic reasoning} frames a task as choosing the output that is most probable given the input. Such reasoning would be influenced by the prior probability of the output: a probabilistic reasoner should show accuracy that increases as the prior probability of the correct answer increases.
\end{enumerate}

\begin{figure}
    \centering
    \includegraphics[width=.5\textwidth]{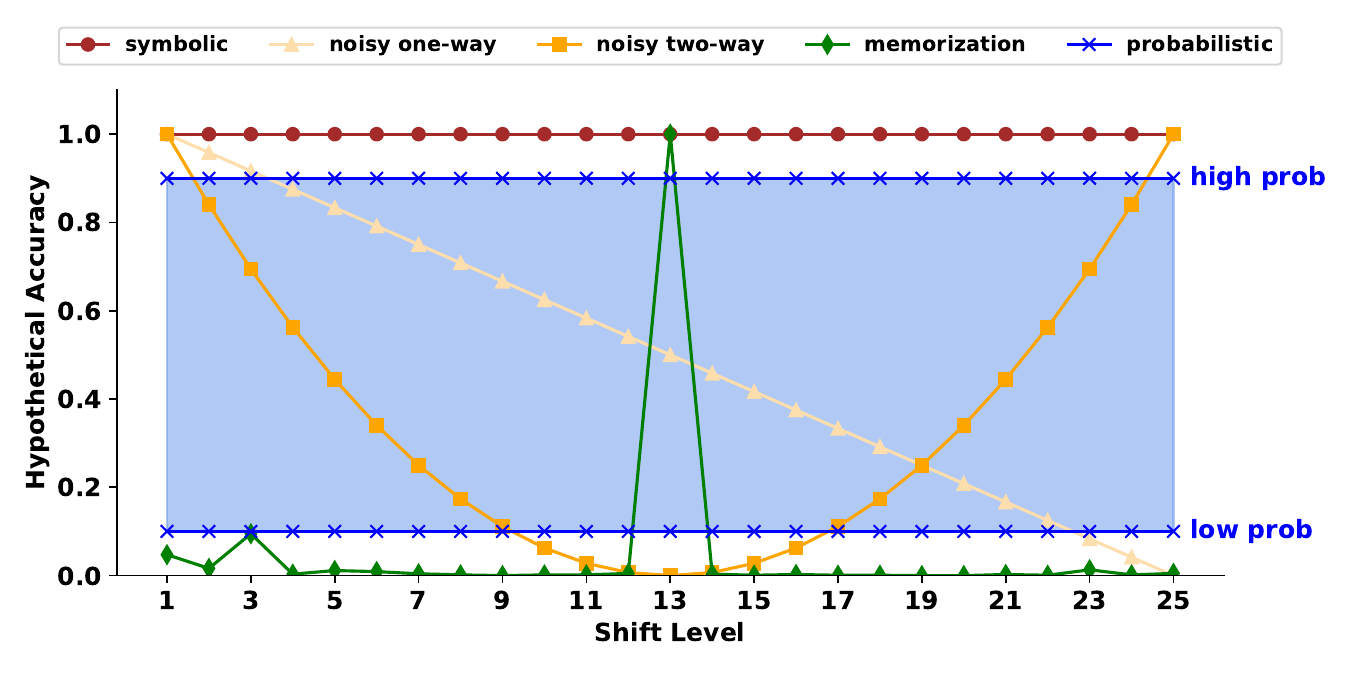}
        \caption{\textbf{Hypothetical} accuracy vs. shift-level for various types of reasoning. Under \textcolor{lightorange}{noisy one-way}, the model only shifts letters backward; under \textcolor{orange}{noisy two-way}, it adopts the shorter path between going forward and backward. The hypothetical \textcolor{darkgreen}{memorization} accuracy is based on shift level frequencies in internet corpora. \textcolor{blue}{Probabilistic} would involve much higher scores on \textcolor{blue}{high prob} than \textcolor{blue}{low prob}.}
        \label{fig:reasoning}
\end{figure}

\noindent
\autoref{fig:reasoning} illustrates the hypothetical performance trends that would be observed in a system adopting each reasoning approach. These approaches are not mutually exclusive; e.g., a reasoner could be influenced by both probability and memorization. Indeed, as discussed below, we find that LLMs' performance when prompted with CoT displays hallmarks of several different types of reasoning; see \autoref{fig:preview}, panel 4 for GPT-4 and \autoref{fig:claude_llama_acc} in the Appendix for Claude 3 and Llama 3.1.

First, accuracy generally decreases as the shift level increases, a hallmark of noisy reasoning. LLMs' performance is, in particular, indicative of a two-way version of noisy reasoning in which it can decode a message by shifting letters forward or backward (e.g., instead of decoding a shift of 25 by shifting back 25 letters, it could instead shift forward 1 letter, as doing so requires fewer steps); this two-way nature shows up in the way that accuracy increases as the shift level changes from 20 to 25.

Second, evidence of probabilistic reasoning can be seen in the fact that accuracy is substantially higher for high prob (the highest-probability bin, \texttt{bin 1}) than low prob (the lowest-probability bin, \texttt{bin 5}). High prob / low prob refer to the probability of the words that are the correct answers when our examples are decoded, where probability is quantified using GPT-2 as described in Section \ref{ss:word_dataset}. The \q{high prob} cases are common words such as \textit{\{'mariner', 'shrines', 'paywall', \dots\}}, while the \q{low prob} cases are nonsense letter sequences such as \textit{\{'xcbrouw', 'jsxrouw', 'levjspx', \dots\}}.

Finally, although a shift level of 13 requires the most reasoning steps of any shift level (assuming decoding can be done forward or backward), there is a spike in accuracy at a shift level of 13. As discussed above, this spike is a hallmark of memorization since 13 is the most common shift level in natural corpora.

For the upcoming detailed analysis experiments, we use GPT-4 as the main reference model, since Claude 3 and Llama 3.1 exhibited similar trends as GPT-4 on the evaluations presented so far.\footnote{The detailed results of \texttt{Llama-3.1-405B-Instruct} and \texttt{claude-3-opus-20240229} are shown in Appendix \S\ref{appx:new_models}.}

\subsection{A simple probabilistic approach to modeling the reasoning process}
\label{ss:cot_factors}

\label{sec:method_lr_model}
To make these intuitively-stated observations more rigorous, we perform a logistic regression to determine the statistical significance of several factors. 
The outcome variable is a binary variable indicating whether GPT-4 got the correct answer on each example. We include the following predictors: 
\begin{itemize}[noitemsep,topsep=0pt,leftmargin=1.5em]
    \item $\mathbf{input\_logprob}$: log probability of the encoded input text as measured by GPT-2 \cite{radford2019language}. The inputs tend to have a very low probability because they are enciphered.  
    \item $\mathbf{output\_logprob}$: log probability of the ground-truth output text as measured by GPT-2. 
    \item $\mathbf{shift\_freq}$: we used the frequency of occurrence of all shift levels that \citet{mccoy2023embers} provided based on analysis of the C4 corpus \cite{raffel2020exploring}. The assumption is that the distribution of shifts in C4 is similar to the distribution in the training data for GPT-4.
    \item $\mathbf{shift\_level}$: the number of steps that must be performed to decode each letter; this feature is added to account for one-way reasoning.
    \item $\bm{\min}\mathbf{(shift\_level, 26 - shift\_level)}$: this value is the minimum number of steps that must be performed to decode each letter, under the assumption that decoding can be done by moving $shift\_level$ steps backward or $(26 - shift\_level)$ steps forward; as discussed above, GPT-4 indeed shows evidence of using both of these decoding directions.
\end{itemize}


\noindent
Several of these variables correspond to the critical properties that are indicative of our hypothesized reasoning processes: $output\_logprob$ should have a significant effect if probabilistic reasoning is used, $shift\_freq$ should have a significant effect if memorization is used, and $\min (shift\_level,26-shift\_level)$ quantifies the difficulty of the task, which should have a significant effect if noisy reasoning is used. The remaining factors are included as potential confounds to control for.
The overall logistic regression thus took the following form:
\begin{equation*}
\label{eq:logistic_reg}
\begin{aligned}
& correct   \sim  input\_logprob  + output\_logprob  \\
& \qquad\qquad + shift\_freq + shift\_level \\
& \qquad\qquad + \min (shift\_level,26-shift\_level) 
\end{aligned}
\end{equation*}

\paragraph{Logistic regression results.  }
The following features had a statistically significant effect on model performance: $output\_logprob$, $shift\_freq$, $shift\_level$, and $\min (shift\_level,26-shift\_level)$ 
($p < 10^{-15}$ in all cases).
These results therefore quantitatively support the conclusion that GPT-4 incorporates processes based on probability, memorization, and noisy reasoning (both forward and backward).
\begin{figure}
\centering
\includegraphics[width=0.5\textwidth]{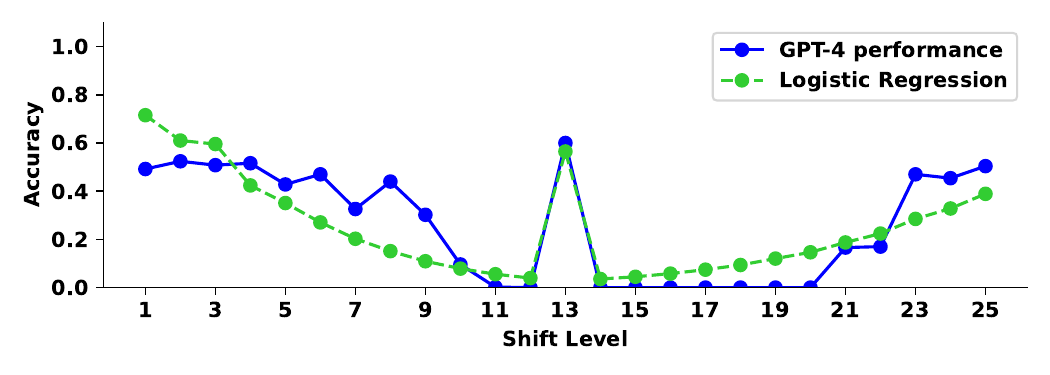}
    \caption{
The \textcolor{green!50!black}{logistic regression curve} captures the overall trend exhibited by \textcolor{blue}{GPT-4}.}
    \label{fig:lr_plot}
\end{figure}

\autoref{fig:lr_plot} shows the predictions of the logistic regression compared to GPT-4's actual performance. 
The logistic regression correctly predicts the main trends in GPT-4's behavior (as expected, it does not match the curve exactly due to the simplicity of the model). 
In the next few subsections, we conduct some additional experiments that investigate each hypothesized reasoning type in more detail.

\begin{figure}
\centering
    \includegraphics[width=.5\textwidth]{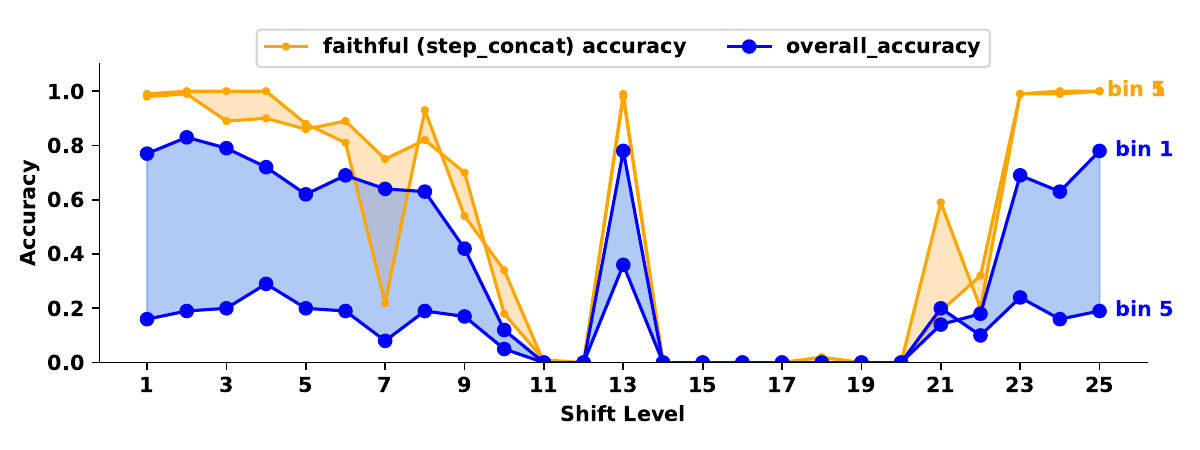}
        \caption{\textbf{Actual} \textcolor{blue}{overall decoding accuracy} vs. \textcolor{orange}{\textit{faithful} accuracy} across shift levels showing the effects of probability. The effect is amplified for low probability outputs as seen in the larger drop in accuracy between the orange and blue \texttt{bin 5} (low probability) lines.}
\label{fig:concat_step_acc_vs_oeverall_acc}
\end{figure}

\subsection{Analyzing the effect of probability}
\label{ss:prior_effects}



If an LLM is influenced by probability, we would expect to occasionally observe \textit{unfaithfulness} between the chain of reasoning steps produced by the LLM and the LLM's final answer. Specifically, if the individual reasoning steps would point to a final output that is low-probability, a probabilistic reasoner might instead produce a different final answer that has a higher probability. For example, in our CoT experiments, each step produces one letter, and these letters must be concatenated to form the final answer. If the individual step outputs are \textit{S, T, A, Z}, the final answer should be \textit{STAZ}, but a model might instead ``self-correct'' by producing the higher-probability word \textit{STAY}. 

Such unfaithfulness can help or hurt the model. When the correct answer truly is a low-probability word such as \textit{STAZ}, then correcting to \textit{STAY} would reduce the model's accuracy. However, if the model had made a mistake during the reasoning chain---such as by producing \textit{S, T, A, Z} when the chain should have been \textit{S, T, A, Y}---then correcting to \textit{STAY} would rescue the model from its error.

To investigate unfaithfulness, we compare the \textit{faithful accuracy} that would be obtained by concatenating GPT-4's step outputs to the actual overall accuracy. 
We indeed observe that overall accuracy is generally lower than faithful accuracy, illustrating that unfaithfulness occurs. 
Further, the drop in accuracy is more pronounced in the low-probability setting than the high-probability setting, which is consistent with the intuition that the lower the probability of a concatenated answer is, the more likely it will be that a probability-reliant model will be steered away from that answer. See \autoref{fig:concat_step_acc_vs_oeverall_acc} 
for the full results.


\autoref{tab:combined-unfaithfulness} provides a more detailed view of unfaithfulness. 
Incorrect intermediate chains (i.e., concatenated step outputs) are followed by correct final answers much more often in the setting where the correct answer has a high probability (\textcolor{ForestGreen}{34\%} and \textcolor{ForestGreen}{55\%} of the time for rot-4 and rot-13, respectively) than in the low probability setting (\textcolor{red}{1\%} and \textcolor{red}{19\%} of the time for rot-4 and rot-13, respectively). On the other hand, correct intermediate chains are followed by incorrect final answers \textit{less} often in the high probability setting (\textcolor{ForestGreen}{7\%} and \textcolor{ForestGreen}{1\%} of the time for rot-4 and rot-13, respectively) than in the low-probability setting (\textcolor{red}{14\%} and \textcolor{red}{9\% }of the time for rot-4 and rot-13, respectively).

These results support the hypothesis that GPT-4 over-relies on the prior probability of potential outputs (see \citet{jang2023can} for some related observations).
If the answer has a high probability of occurrence, the model's priors favor generating it even if its intermediate reasoning steps suggest an alternative output. Conversely, if the answer is of lower probability, then even if the chain of reasoning is correct, the priors exert a detrimental influence leading to incorrect final answers.

\begin{table}
  \centering
  \footnotesize 
  \begin{adjustbox}{max width=\columnwidth} 
    \begin{tabular}{c|c|cc|cc}
      \toprule
      \multirow{2}{*}{\textbf{Prob}} & \multirow{2}{*}{\textbf{\begin{tabular}[c]{@{}c@{}}Chain Steps\\ Output\end{tabular}}} & \multicolumn{2}{c|}{\textbf{\texttt{rot-4}}} & \multicolumn{2}{c}{\textbf{\texttt{rot-13}}} \\
      \cline{3-6}
      & & Correct & Incorrect & Correct & Incorrect \\
      \midrule
      \multirow{2}{*}{High} & Correct & 19 & \textcolor{ForestGreen}{7} & 15 & \textcolor{ForestGreen}{1} \\
      & Incorrect & \textcolor{ForestGreen}{34} & 40 & \textcolor{ForestGreen}{55} & 29 \\
      \hline
      \multirow{2}{*}{Low} & Correct & 7 & \textcolor{red}{14} & 7 & \textcolor{red}{9} \\
      & Incorrect & \textcolor{red}{1} & 78 & \textcolor{red}{19} & 65 \\
      \bottomrule
    \end{tabular}
  \end{adjustbox}
  \caption{Confusion matrices (100 examples; 2 probability bins \{high, low\}) for \texttt{rot-4} and \texttt{rot-13}. \textit{Effect of memorization}: incorrect step outputs lead to correct final answers more often for \texttt{rot-13} than \texttt{rot-4}. \textit{Effect of probability}: Unfaithfulness has a positive effect more often in \textcolor{ForestGreen}{high-probability bins} than in \textcolor{red}{low-probability bins}; Conversely, unfaithfulness has a negative effect more often in \textcolor{red}{low-probability bins} than in \textcolor{ForestGreen}{high-probability bins}.}
  \label{tab:combined-unfaithfulness}
\end{table}


\subsection{Analyzing the effects of noise}
\label{ss:noise_effects}
\begin{figure}
\centering
\includegraphics[width=0.5\textwidth]{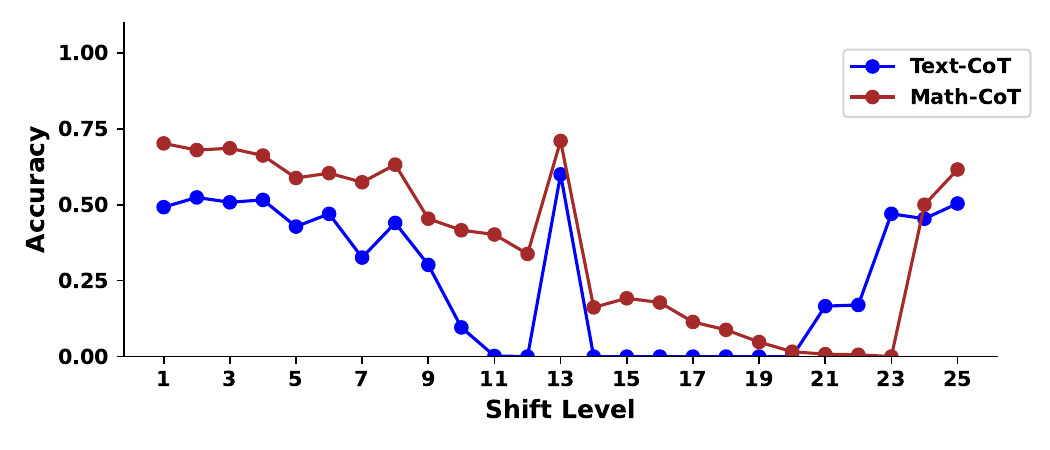}
    \caption{Accuracy for Text- and Math- CoT prompt styles with GPT-4. Math-CoT performs better than Text-CoT, but both display evidence of memorization as accuracy is highest for shift level 13---the most frequent shift in real-world corpora.}
    \label{fig:text_vs_math_cot}
\end{figure}

The statistically significant impact of $shift\_level$ is evidence that GPT-4's CoT behavior is in part a noisy version of symbolic reasoning. 
Accuracy falls as the shift level increases from 1 to 12 and then recovers at higher shift levels 
(\autoref{fig:text_vs_math_cot}), consistent with a noisy reasoning process in which deciphering each letter with a shift level of $n$ involves $\min (n,26-n)$ implicit steps, with noise that gives each step some probability of being performed incorrectly. Note that the implicit steps referred to here are different from the steps that are explicitly produced in the chain of thought: the chain of thought uses one explicit step per letter (\autoref{fig:text_cot_prompt}), but here we are discussing the operations that must be implicitly carried out within each step of this chain in order to decode each letter.

\begin{figure*}
    \centering
\includegraphics[width=0.85\textwidth]{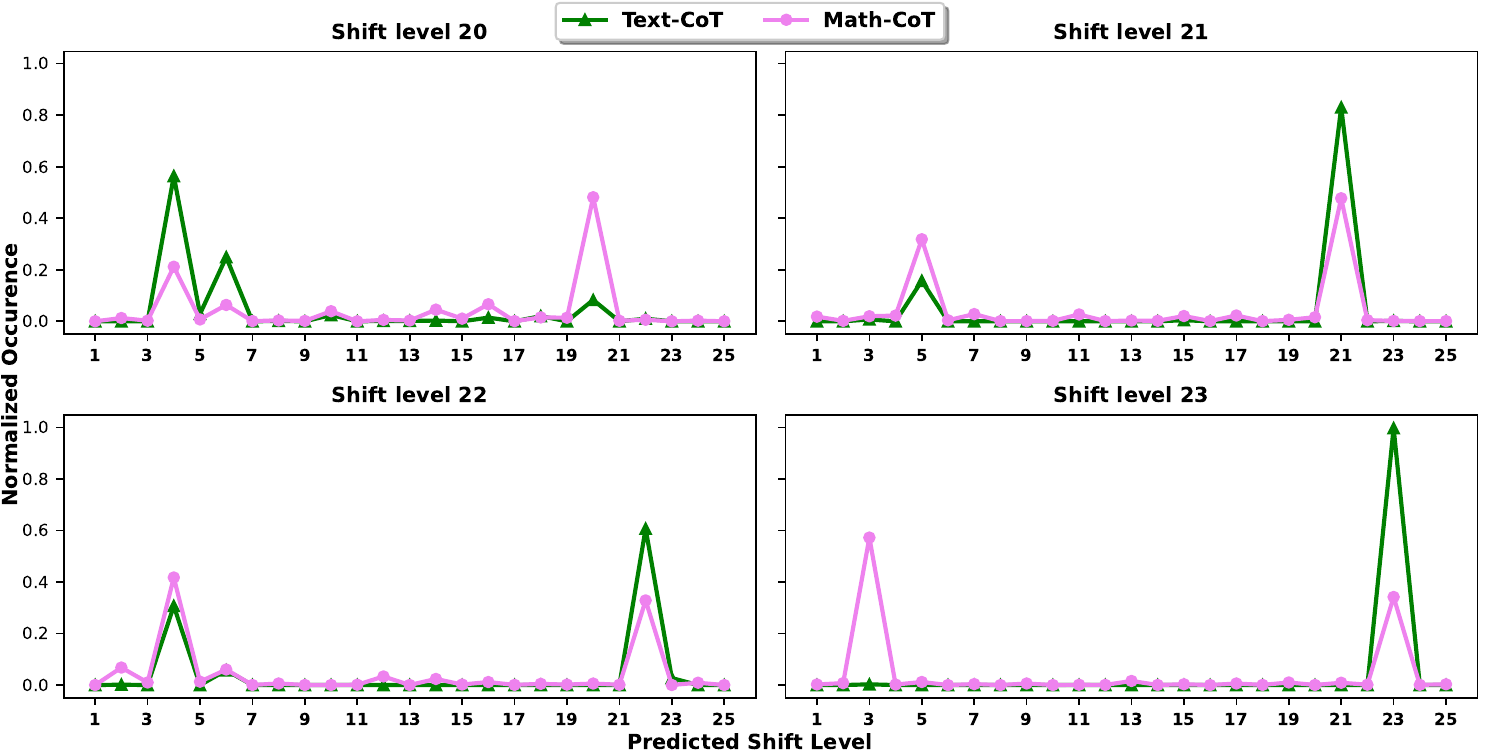}
    \caption{Normalized frequency distribution vs. predicted $shift\_level$ of step answers for \texttt{rot-20} to \texttt{rot-23}. The appearance of peaks at $26 - shift\_level$ in \textcolor{pink}{Math-CoT} and \textcolor{green!50!black}{Text-CoT} prompts showcases the model's noisy attempt in taking the shorter path---i.e., moving $26-x$ shifts forward. 
    }
    \label{fig:text_math_plot2_20-23}
\end{figure*}

\paragraph{Noise relating to complementary shift levels. } 
We have argued that the relation between accuracy and shift level is evidence that GPT-4 uses a two-way strategy. That is, accuracy is high for small shifts such as 1 but also for large shifts such as 25, which could plausibly be explained by GPT-4 implicitly selecting whichever direction will minimize the number of steps it needs to compute---shifting letters backward for small shift levels or forward for large shift levels.
This two-way strategy is effective in that it supports strong performance on large shift levels such as 25. 
However, we also observe evidence that it contributes to the noise that causes accuracy to decline as the shift level increases.
\autoref{fig:text_math_plot2_20-23} shows the actual shift level that GPT-4 produces for each letter in its chain of thought, for each of four intended shift levels. Across all four of these cases, GPT-4 shows peaks at both $shift\_level$ and $26 - shift\_level$.
Thus, while some of the noise affecting the reasoning process may be random, it appears that at least some of the noise can be attributed to confusion between possible shift levels. Decoding a shift level of $n$ can be done by shifting backward $n$ steps or forward $26-n$ steps, but it appears that GPT-4 sometimes mixes up these two strategies by shifting forward $n$ steps (or, equivalently, shifting backward $26-n$ steps), contributing to the overall noise.

\paragraph{Discretization.}
Another factor that interacts with noise is \texttt{temperature}. In principle, if CoT entailed pure symbolic reasoning, it would assign 100\% probability to the correct continuation (i.e., each predicted next token) and 0\% probability to everything else. If so, the temperature used would not affect performance. 
However, we observe that CoT scores better with a low temperature. For instance, in \texttt{rot-13}, GPT-4's accuracy is 30.0\% at \texttt{temperature=0} and 0.33\% at \texttt{temperature=1}. This shows that its predicted distribution over the vocabulary is not fully discrete---it has some noise in it that a low temperature can remove. However, even a temperature of 0 does not make the performance perfect because noise does not solely arise in the final distribution over the vocabulary (which temperature modifies) but also influences the implicit intermediate steps used to produce that distribution (which temperature does not change).

\subsection{Analyzing the effect of memorization}
\label{ss:memorization_effects}
To further investigate memorization, we focus on \texttt{rot-13}, because frequency is generally confounded with simplicity for the other shift levels (e.g.,  \texttt{rot-1} is simple as well as frequent).
$13$ is the most frequent shift level \cite{mccoy2023embers}, and we observe in 
\autoref{fig:text_vs_math_cot}
that GPT-4 shows a spike in accuracy at this shift level in both Text-CoT and Math-CoT, providing strong evidence that memorization plays a role in GPT-4's CoT performance. 

We also observe that memorization influences \textit{unfaithfulness}. Consider the cells in \autoref{tab:combined-unfaithfulness} that involve incorrect chain steps outputs but correct final answers; such cases are much more common for \texttt{rot-13} than for other levels, including \texttt{rot-4} (the other shift level shown in that table); e.g., in the high-probability case, 55\% of rot-13 examples fall in this category, while only 34\% of rot-4 examples do. 
This pattern also provides some evidence for memorization: for \texttt{rot-13}, the model may have two \q{paths} for producing the final output---it could use the chain of thought it has produced, or it could go directly from the input to the output due to memorization. Thus, when it produces the final output, it might implicitly weigh both of those paths, which helps it to correct faulty chains because it has a back-up path to consider. However, for \texttt{rot-4}, it may be that only the path involving the chain of thought is available, such that GPT-4 cannot fix incorrect chains as easily because it does not have this alternative path to fall back on.



\subsection{The role of intermediate reasoning steps}
\label{ss:hidden_reasoning}


Finally, we study the role of the intermediate reasoning steps that are involved with CoT prompting---both the chain that GPT-4 produces and the chain provided in the demonstration.

\paragraph{Strong reliance on the surface strings produced in reasoning steps. }
First we focus on the chain of thought that GPT-4 produces before providing its final answer. We consider two potential roles that this chain could have. First, it could be that the chain is helpful because it provides text that is useful for GPT-4 to condition on in later steps. Alternatively, it could be that the critical aspect of CoT reasoning is internal---rather than depending on the text that is produced, CoT could be helpful because it gives the LLM the opportunity to internally perform additional reasoning steps.

To disentangle these possibilities, we modify the prompt so that GPT-4 is told to perform the same steps of reasoning as before, but to have the intermediate output that it produces be uninformative. Specifically, we used the Text-CoT prompt but instructed the model to \textit{not reveal step answers and instead output a *}. The step answers in the demonstration were also replaced by `*'; thus we left the format of reasoning intact but the expected generation token was no longer a component of the final answer.
Next, we asked the model to explicitly \textit{think about the correct letter that should go in the place of the * but just not write it down}. In the demonstration, we first provided an example with all step answers and then repeated the same example but with a * in place of each output letter (see Appendix \ref{appx:experiments:hidden_prompt} \autoref{fig:hidden_prompts} for prompts). 

In both settings, performance 
was similar to that of the standard prompting variant (shown in \autoref{fig:preview}, panel 2). This is evidence that CoT depends on ``self-conditioning''---explicitly producing text that will be useful as context to condition on when producing the final answer. Merely instructing a model to \q{think} silently is not helpful, suggesting that the reasoning does not occur internally. 
These results corroborate prior work finding that CoT is unhelpful when the model is told to produce contentless ``filler'' tokens instead of contentful text
\citep{lanham2023measuring}; models can reason internally when explicitly trained to do so
\citep{pfau2024let}, but current LLMs without this explicit training do not seem to have this ability.

\paragraph{Little reliance on the validity of the demonstration.}

In experiments described in Appendix \ref{appx:experiments:hidden_prompt}, we also find that the validity of the reasoning shown in the prompt does not have a strong effect of CoT performance for shift ciphers. That is, even when the demonstration is perturbed such that it contains many errors, GPT-4's CoT performance remains approximately the same. This finding corroborates prior work showing that the validity of demonstrations did not matter much \cite{wang-etal-2023-towards, madaan2022text, ye-etal-2023-complementary}; the demonstration seems to merely guide the model to solve the task by providing a format to generate accurate reasoning steps \cite{min2022rethinking}. 

\section{Conclusion}
\label{sec:conclusion}
We have used the case study of shift ciphers to disentangle the factors that influence CoT reasoning, with a focus on characterizing what \textit{type} of reasoning is used in models prompted with CoT. We found that CoT performance is statistically significantly influenced by the probability of occurrence of the expected task output, the frequency of the task in corpora, and the number of reasoning steps that must be (implicitly) performed. These results suggest that CoT reasoning can be characterized as probabilistic, memorization-influenced noisy reasoning, meaning that LLM behavior displays traits of both memorization and generalization. 

\section*{Limitations}
We have conducted extensive studies on a single task, decoding shift ciphers; we chose this task because it enables us to separate memorization from reasoning in a controlled manner as explained in \autoref{ss:motivation}, and these factors cannot be easily disentangled for most other tasks. Our prompts contain one example demonstration (i.e. one-shot CoT prompting), but  a single demonstration contains multiple reasoning steps which provide more than just one reference of decoding to the model. 
In addition, while our experiments showed that these frontier models display some hallmarks of true reasoning, they also showed some ways in which they make errors due to a reliance on memorization and probability; future work could investigate how these limitations could be overcome.

\section*{Ethical Considerations}
We do not believe that this work raises major potential risks.  As is the case with any analysis work, there is some risk that our results could lead some readers to overestimate or underestimate the abilities of LLMs, either of which could have negative consequences: overestimation can contribute to hype, while underestimation can result in the field paying too little attention to potential harms. However, we believe that this risk is minimal because we have aimed to present our results in a balanced way that highlights both strengths and limitations.


\section*{Acknowledgements}

This work was supported in part by the National
Science Foundation SBE Postdoctoral Research Fellowship under Grant No.\ 2204152. Any opinions, findings, and conclusions or recommendations expressed in this material are those of the authors and do not necessarily reflect the views of the National Science Foundation.

\bibliography{custom}

\appendix
\section{Additional Details \& Experiments}


\subsection{Math- \& Number- CoT prompts}
\label{appx:prompts}

While the Standard prompt (\autoref{fig:standard_prompt}) yields poor performance on most shift levels, the Number-CoT prompt (\autoref{fig:math_only_prompt}) in contrast gives nearly perfect scores across the shift levels. It is to be noted that some miscellaneous noise is captured in the Number-CoT case. This arises mostly due to incomplete generations/half-completed chains requiring more tokens than needed as the model does some additional sub-reasoning steps, and in very rare cases produces numbers greater than 25.
The Math-CoT prompt used is shown in  \autoref{fig:math_cot_prompt}. 
\begin{figure}
\centering
\begin{tcolorbox}[
    width=0.5\textwidth,
    fontupper=\ttfamily,
    left=0.5mm,
    right=0.5mm,
    top=0.5mm,
    bottom=0.5mm,
    title=Math-CoT,
    center title,
    fonttitle=\tiny\ttfamily]
\tiny
\textcolor{blue}{Rot-13 is a cipher in which each letter is shifted 13 positions forward in the alphabet.} \textcolor{orange}{For example, here is a message written in rot-13:\\
Rot-13 text: "fgnl"}\\

\textcolor{orange}{To decode this message, we need to shift each letter 13 positions backward. Let’s start by writing the position-letter mapping for the alphabet:}

\textcolor{orange}{a -> 0\\
b -> 1\\
$\cdots$}\\

\textcolor{orange}{Next, we find the encoded letter as follows: \\
Position of original letter = (Position of given letter - 13) mod 26} \\

\textcolor{orange}{Then map the found position to the corresponding letter using the letter-position mapping.}\\

\textcolor{orange}{Using this,}\\
\textcolor{purple}{1. f -> (5 - 13) mod 26 = 18 -> s\\
2. g -> (6 - 13) mod 26 = 19 -> t\\
3. n -> (13 - 13) mod 26 = 0 -> a\\
4. l -> (11 - 13) mod 26 = 24 -> y}\\
\textcolor{orange}{Therefore, the original text is: "stay"}\\

Here is another message in rot-13. Decode this message one letter at a time. On the last line, write the words "Original text:" followed by the decoded message:\\
Rot-13 text: <test\_input>
\end{tcolorbox}
\caption{Math-CoT prompt.}
\label{fig:math_cot_prompt}
\end{figure}

\begin{figure}
\centering
\begin{tcolorbox}[
    width=0.5\textwidth,
    fontupper=\ttfamily,
    left=0.5mm,
    right=0.5mm,
    top=0.5mm,
    bottom=0.5mm,
    title=Number-CoT,
    center title,
    fonttitle=\tiny\ttfamily]
\tiny
\textcolor{blue}{Shift-13 is a process in which each number is shifted 13 positions forward until it reaches 26 and subsequently circles back to 1.} \textcolor{orange}{For example, here is a sequence of numbers written in shift-13:\\
shift-13 sequence: "5,6,13,11"}\\

\textcolor{orange}{To decode this sequence, we need to shift each number 13 positions backward.}\\
\textcolor{orange}{New position = (Given position - 13) mod 26}\\

\textcolor{orange}{Using this,}\\
\textcolor{purple}{1. 5 -> (5 - 13) mod 26 -> 18\\
2. 6 -> (6 - 13) mod 26 -> 19\\
3. 13 -> (13 - 13) mod 26 -> 0\\
4. 11 -> (11 - 13) mod 26 -> 24}\\

\textcolor{orange}{Therefore, the original sequence of numbers is: "18,19,0,24"}\\

Here is another sequence of numbers in shift-13. Decode this sequence one number at a time. On the last line, write the words "Original sequence:" followed by the decoded sequence:\\
shift-13 sequence: <test\_input>
\end{tcolorbox}
\caption{The shift cipher task reformatted to be performed in the domain of numbers rather than the domain of letters (Number-CoT). Here both the encoded input and decoded output are number sequences abstracting out the alphabet and its influences.}
\label{fig:math_only_prompt}
\end{figure}

\begin{figure*}
\centering
    \includegraphics[width=\textwidth]{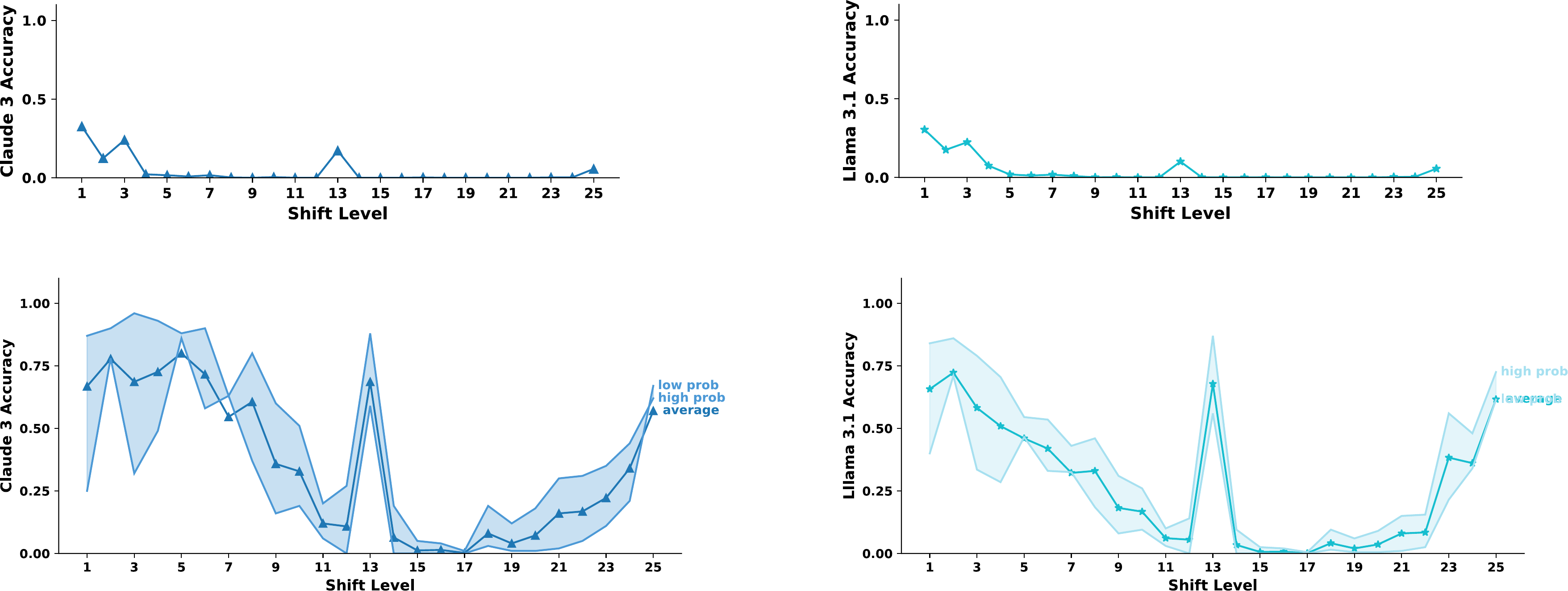}
        \caption{Accuracies with Claude 3 and Llama 3.1. The top panel shows the performance when prompted with Standard prompting. The bottom panel shows the trend with Text-CoT. The shaded regions indicate the gap between the low and high probability bins.}
\label{fig:claude_llama_acc}
\end{figure*}

\subsection{Results with Llama 3.1 and Claude 3}
\label{appx:new_models}
We display the results with Llama 3.1 and Claude 3 when prompted with standard prompts and Text-CoT in \autoref{fig:claude_llama_acc}. Interestingly, Llama 3.1 was trained on CoT data, such that even when it is prompted with standard prompts it generates reasoning steps. However, as we can see from \autoref{fig:claude_llama_acc}, with Text-CoT there is a further enhancement in scores across the shift levels. 
Overall, both models display trends similar to that of GPT-4 shown in \autoref{fig:preview}.

\subsection{Impact of producing explicit reasoning steps and validity of steps in demonstration.}
\label{appx:experiments:hidden_prompt}
\begin{figure*}
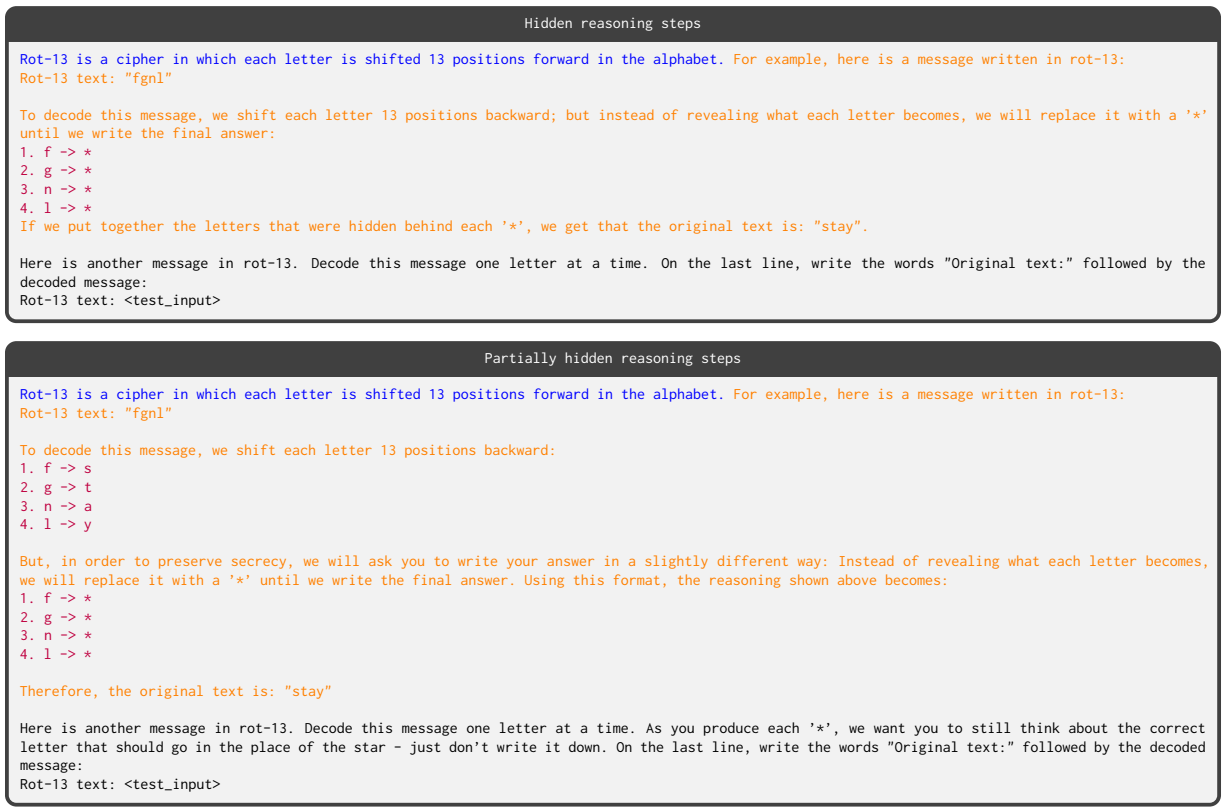

\begin{tcolorbox}[
    width=\textwidth,
    fontupper=\ttfamily,
    left=0.5mm,
    right=0.5mm,
    top=0.5mm,
    bottom=0.5mm,
    title=Hidden reasoning steps,
    center title,
    fonttitle=\tiny\ttfamily]
\tiny
\textcolor{blue}{Rot-13 is a cipher in which each letter is shifted 13 positions forward in the alphabet.} \textcolor{orange}{For example, here is a message written in rot-13:\\
Rot-13 text: "fgnl"}\\

\textcolor{orange}{To decode this message, we shift each letter 13 positions backward; but instead of revealing what each letter becomes, we will replace it with a '*' until we write the final answer:}\\
\textcolor{purple}{1. f -> *\\
2. g -> *\\
3. n -> *\\
4. l -> *}\\
\textcolor{orange}{If we put together the letters that were hidden behind each '*', we get that the original text is: "stay".}\\

Here is another message in rot-13. Decode this message one letter at a time. On the last line, write the words "Original text:" followed by the decoded message:\\
Rot-13 text: <test\_input>
\end{tcolorbox}
\begin{tcolorbox}[
    width=\textwidth,
    fontupper=\ttfamily,
    left=0.5mm,
    right=0.5mm,
    top=0.5mm,
    bottom=0.5mm,
    title=Partially hidden reasoning steps,
    center title,
    fonttitle=\tiny\ttfamily]
\tiny
\textcolor{blue}{Rot-13 is a cipher in which each letter is shifted 13 positions forward in the alphabet.} \textcolor{orange}{For example, here is a message written in rot-13:\\
Rot-13 text: "fgnl"}\\

\textcolor{orange}{To decode this message, we shift each letter 13 positions backward:}\\
\textcolor{purple}{1. f -> s\\
2. g -> t\\
3. n -> a\\
4. l -> y}\\

\textcolor{orange}{But, in order to preserve secrecy, we will ask you to write your answer in a slightly different way: Instead of revealing what each letter becomes, we will replace it with a '*' until we write the final answer. Using this format, the reasoning shown above becomes:}\\
\textcolor{purple}{1. f -> *\\
2. g -> *\\
3. n -> *\\
4. l -> *\\
}\\
\textcolor{orange}{Therefore, the original text is: "stay"}\\

Here is another message in rot-13. Decode this message one letter at a time. As you produce each '*', we want you to still think about the correct letter that should go in the place of the star - just don't write it down. On the last line, write the words "Original text:" followed by the decoded message:\\
Rot-13 text: <test\_input>
\end{tcolorbox}

\caption{Text-CoT prompt consisting of hidden (top) and partially hidden (bottom) \textcolor{purple}{reasoning steps}.}
\label{fig:hidden_prompts}
\end{figure*}

\autoref{fig:hidden_prompts} shows the prompts used to test the importance of producing explicit reasoning steps by forcing the model to \q{think} silently.

\paragraph{Outputs from \textit{silent thinking.}}
We observe that many of the outputs produced in the case when the reasoning step answers are * are related to the task of shift ciphers. It seems that the model now relies on terms such as \textit{cipher} and \textit{decode} that are present in the description, perhaps relying less on the reasoning steps because these reasoning steps are now less informative given that each step's output is now *. Influenced by the probabilistic effects discussed before it produces outputs having a high probability and containing words related to security, safety, and programming such as \textit{encryption},  \textit{code cracker}, \textit{decoded}, \textit{Javascript}, and \textit{Instagram}.

\paragraph{Validity.}
Next we consider the chain that is provided to GPT-4 within the prompt. Prior work has found that the validity of the reasoning shown in the prompt does not have a strong effect on CoT performance; CoT is approximately as helpful when the chain of thought in the prompt contains errors as when it does not \cite{wang-etal-2023-towards, madaan2022text, ye-etal-2023-complementary}. Here we test whether this observation extends to shift ciphers.

\begin{figure*}
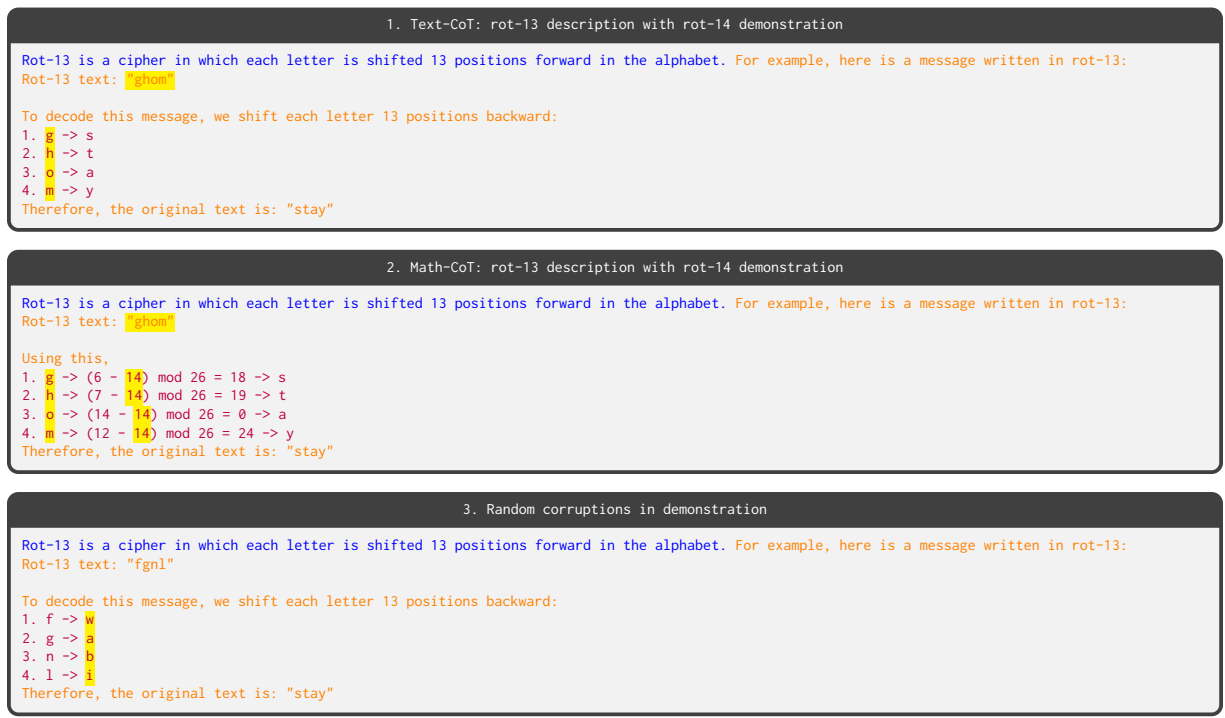

\centering
\begin{tcolorbox}[
    width=\textwidth,
    fontupper=\ttfamily,
    left=0.5mm,
    right=0.5mm,
    top=0.5mm,
    bottom=0.5mm,
    title=1. Text-CoT: rot-13 description with rot-14 demonstration,
    center title,
    fonttitle=\tiny\ttfamily]
\tiny
\textcolor{blue}{Rot-13 is a cipher in which each letter is shifted 13 positions forward in the alphabet.} \textcolor{orange}{For example, here is a message written in rot-13:\\
Rot-13 text: \hl{"ghom"}}\\

\textcolor{orange}{To decode this message, we shift each letter 13 positions backward:}\\
\textcolor{purple}{1. \hl{g} -> s\\
2. \hl{h} -> t\\
3. \hl{o} -> a\\
4. \hl{m} -> y}\\
\textcolor{orange}{Therefore, the original text is: "stay"}
\end{tcolorbox}
        \centering
\begin{tcolorbox}[
    width=\textwidth,
    fontupper=\ttfamily,
    left=0.5mm,
    right=0.5mm,
    top=0.5mm,
    bottom=0.5mm,
    title=2. Math-CoT: rot-13 description with rot-14 demonstration,
    center title,
    fonttitle=\tiny\ttfamily]
\tiny
\textcolor{blue}{Rot-13 is a cipher in which each letter is shifted 13 positions forward in the alphabet.} \textcolor{orange}{For example, here is a message written in rot-13:\\
Rot-13 text: \hl{"ghom"}}\\

\textcolor{orange}{Using this,}\\
\textcolor{purple}{1. \hl{g} -> (6 - \hl{14}) mod 26 = 18 -> s\\
2. \hl{h} -> (7 - \hl{14}) mod 26 = 19 -> t\\
3. \hl{o} -> (14 - \hl{14}) mod 26 = 0 -> a\\
4. \hl{m} -> (12 - \hl{14}) mod 26 = 24 -> y}\\
\textcolor{orange}{Therefore, the original text is: "stay"}
\end{tcolorbox}

        \centering
\begin{tcolorbox}[
    width=\textwidth,
    fontupper=\ttfamily,
    left=0.5mm,
    right=0.5mm,
    top=0.5mm,
    bottom=0.5mm,
    title=3. Random corruptions in demonstration,
    center title,
fonttitle=\tiny\ttfamily]
\tiny
\textcolor{blue}{Rot-13 is a cipher in which each letter is shifted 13 positions forward in the alphabet.} \textcolor{orange}{For example, here is a message written in rot-13:\\
Rot-13 text: "fgnl"}\\

\textcolor{orange}{To decode this message, we shift each letter 13 positions backward:}\\
\textcolor{purple}{1. f -> \hl{w}\\
2. g -> \hl{a}\\
3. n -> \hl{b}\\
4. l -> \hl{i}}\\
\textcolor{orange}{Therefore, the original text is: "stay"}
\end{tcolorbox}

\caption{Prompt snippets with different types of mismatches/corruptions. The corruptions that were introduced are highlighted in yellow.} 
\label{fig:desc_demo_prompt}
\end{figure*}

We first tried applying random corruptions to the steps in the demonstration.
Each step answer in the demonstration was replaced by a random letter; the format of the chain was left unchanged, with only the produced letters being modified.
We also tried applying systematic corruptions by having the linguistic description of the task specify
\texttt{rot-13} yet having the demonstration illustrate \texttt{rot-14}.  \autoref{fig:desc_demo_prompt} shows the prompts used.

With both corruptions, GPT-4's performance remained approximately as strong as when it was given accurate, uncorrupted demonstrations.
This result corroborates the aforementioned prior work that found no strong correlation between the validity of the reasoning shown in the demonstration and the model performance. The demonstration seems to merely guide the model to solve the task by providing a format to generate accurate reasoning steps \cite{min2022rethinking}.

\end{document}